# Tailored Patient Information: Some Issues and Questions

**Ehud Reiter**
Dept of Computing Science
University of Aberdeen
Aberdeen AB24 2TN
BRITAIN
ereiter@csd.abdn.ac.uk

**Liesl Osman**
Dept of Medicine and Therapeutics
University of Aberdeen
Aberdeen AB9 2ZD
BRITAIN
l.osman@abdn.ac.uk

**Abstract**

Tailored patient information (TPI) systems are computer programs which produce personalised heath-information material for patients. TPI systems are of growing interest to the natural-language generation (NLG) community; many TPI systems have also been developed in the medical community, usually with mail-merge technology. No matter what technology is used, experience shows that it is not easy to field a TPI system, even if it is shown to be effective in clinical trials. In this paper we discuss some of the difficulties in fielding TPI systems. This is based on our experiences with 2 TPI systems, one for generating asthma-information booklets and one for generating smoking-cessation letters.

## 1 Introduction

Tailored patient information systems are computer programs which generate personalised medical information or advice. There are a growing number of natural-language generation (NLG) projects which fall into this category, such as (Buchanan et al., 1995; Cawsey et al., 1995). There have also been several projects in the medical community which used mail-merge technology to produce personalised medical information, such as (Velicer et al., 1993; Campbell et al., 1994). We are also aware of one mail-merge TPI system that is used commercially (see *http://www.commitedquitters.com*), although it is unclear whether this system is genuinely effective or merely a marketing gimmick.

In this paper, we discuss some of the practical issues and trade-offs involved in building tailored-patient information systems (henceforth referred to as TPI). This discussion is partially based on our experiences in two projects: GRASSIC, a TPI system for asthma booklets which was proven clinically effective but nevertheless never deployed; and a newer project to generate personalised smoking-cessation letters. We believe that the points we raise apply to other TPI systems as well, and perhaps also to other NLG applications.

The rest of this section gives some general background on TPI systems. Section 2 introduces the two specific systems mentioned above. Section 3 discusses some specific issues which affect deployment, and is the heart of the paper. Section 4 gives some concluding comments.

### 1.1 Tailored patient information

TPI systems generate personalised letters, booklets, hypertext, or other text-based documents for patients (or other health-care consumers). Tailoring is based on information about the patient, some of which may be extracted from a standard Patient Record System (PRS) database. TPI systems can be based on NLG or on simpler mail-merge technology; this is an engineering decision (Reiter, 1995), based on what functionality is desired.

TPI systems are usually intended to change either the behaviour or mental state of a patient. For example, TPI systems can be used to help people stop smoking (Velicer et al., 1993); to increase compliance with a treatment regime (Osman et al., 1994); or to reduce anxiety in patients (Cawsey and Grasso, 1996). Usually these goals are stated in clinical terms, and the effectiveness of the TPI system is evaluated in a controlled clinical trial.

## 2 Our Projects

### 2.1 GRASSIC

The GRASSIC system (Osman et al., 1994) used mail-merge techniques to generate personalised asthma-information booklets. Personalisation mainly consisted of making local changes in the document; style and overall document structure were not affected. For example, whenever the booklets discussed medication, they only mentioned the medication prescribed for the target patient; whenever they discussed side-effects, they only mentioned side-effects associated with the prescribed medication; and so forth. An attempt was also made to avoid terminology and names unfamiliar to patients; for example, commercial names were used for medication, instead of scientific names. Although mail-merge technology was used, care was taken to avoid the usual "fill-in-the-blank" form-letter look.

Despite its simple technology, a clinical evaluation showed that GRASSIC was successful in reducing hospital admissions among severe asthma patients. Indeed, severe asthma patients who received the GRASSIC material had half of the number of hospital admission as control patients who received non-personalised material. Thus, GRASSIC improved the health and quality of life of its patients; it also saved the health service approximately £500/patient/year by reducing hospital admissions. These figures imply that if GRASSIC was deployed throughout Scotland, it could save the health service perhaps £5,000,000 per year; deployment throughout the UK might save an order of magnitude more money.

Even though it was clinically effective, however, GRASSIC was never fielded. Instead, when the study was finished the non-personalised booklets were rewritten based on a better understanding of patient needs that was one result of GRASSIC. Also, a single page was added at the front of the booklet where a health professional could (manually) write in the details of a personal management plan for this patient; this required a few minutes at most, and was typically done during a consultation with the patient.

Why was GRASSIC not fielded? Partly this was due to classic technology-transfer issues. For instance, the team which developed GRASSIC was a research group, and did not have the skills and resources necessary to turn the prototype into a fieldable system; this would have required developing better user interfaces, making the code more robust, writing manuals and other supporting documentation, helping users install the system, and so forth. Furthermore, there was no existing development team whose remit covered GRASSIC's functionality, and hence which GRASSIC could naturally be transitioned to.

Another problem was that the developers were concerned that doctors would be reluctant to use GRASSIC, because it was a new technology and did not deliver dramatic and visible benefits to *individual* medical professionals. That is, while fielding GRASSIC might provide significant benefit to the health service as a whole, from the perspective of an individual doctor, who dealt with many kinds of patients in addition to people suffering from severe asthma, the effect of using GRASSIC was a relatively small reduction in the total number of his or her patients admitted to hospital. Given the natural reluctance of many people to adopt new technology, the developers were worried that doctors would in practice be reluctant to learn about and use GRASSIC, even if its use was recommended by the health service.

Because of these problems, the development team decided to go for the alternative approach of improved non-personalised material, plus a limited amount of manual personalisation. No clinical evaluation was done of the alternative approach, but studies elsewhere (such as (Lahdensuo et al., 1996)) have demonstrated the effectiveness of manually-written personal management plans. The manual approach was probably less effective at reducing hospital admissions than the tailored-material produced by GRASSIC, but it could be implemented with the skills and resources available to the development team, and furthermore fit much more naturally into the current working practices of Scottish medical professionals.

It may seem odd, incidentally, to discuss a mail-merge system in a workshop devoted to NLP, but we believe that the fielding/deployment issues that arose with GRASSIC are likely to affect any TPI system, regardless of which technology it is based on.

### 2.2 Smoking-cessation letters

More recently, we have begun working on a TPI system which generates personalised smoking-cessation letters, and which uses some NLG technology (Reiter et al., 1997). Personalisation is based on an "Attitudes Towards Smoking" questionnaire which patients fill out. This project is still at an early stage, but we want to be sure that it can be deployed if it proves effective in clinical trials. Hence we have been trying to develop a better understanding of deployment issues of TPI systems, in the hope that this will help us design the system in an appropriate fashion.

## 3 Deployment Issues

In the course of thinking about why GRASSIC was not fielded, and how to build the smoking-letters system so that it can be fielded, we have identified a number of specific issues. We believe these apply in some manner to all TPI systems, and perhaps to other types of NLG systems as well.

### 3.1 Cost-Effectiveness

Perhaps the most obvious real-world consideration for TPI systems is cost-effectiveness. No one is going to use a TPI system unless it is cheaper than having a person manually write letters, explain patient records, etc. In the medical setting, money will not be spent on TPI unless it is seen as being effective in improving clinical outcomes for patients, and/or saving money for the health service.

We will examine both GRASSIC and our smoking-cessation letters system by this criteria. Incidentally, a general rule of thumb in AI and other advanced computing applications is that such systems need to have a pay-back period of 2-3 years at most, with 1 year being preferable.

If we look at GRASSIC first, there are three comparisons that can be made:
- GRASSIC vs. non-personalised booklets: As pointed out above, GRASSIC has the potential to save the Scottish health service several million pounds per year (assuming that doctors are willing to use the system), which means that its development, fielding, and deployment costs would probably be paid back within a year.
- Manually-tailored vs. non-personalised booklets: We have no data on the effectiveness of the manually-tailored booklets, but our best guess is that they capture most but not all of the benefits of GRASSIC. Since fielding and deployment costs for these booklets are minimal, the pay-back period for using the manually-tailored booklets is very short.
- GRASSIC vs. manually-tailored booklets: With the above assumptions, the pay-back period for GRASSIC compared to the manually-tailored booklets could be more than 3 years.

In short, when compared to the alternative of the manually-tailored letters, GRASSIC may not meet the "pay back within 2-3 years" criteria for cost-effectiveness. A big caveat here, though, is that this assumes that the manually personalised letters are effective at reducing hospital admission rates for severe asthmatics; if this is not the case, than GRASSIC does meet the cost-effectiveness rule.

For the smoking-letters system, it is hard to estimate the monetary value of helping someone quit smoking, but since smoking a pack a day can cut life expectancy by 5 years (Austoker et al., 1994), we would hope that society places a benefit of at least £10,000 on a successful cessation. We do not yet know if our smoking-cessation letters are effective, but if they are successful in convincing 2% of smokers to quit, that will mean a benefit to Scottish society of several hundred million pounds, which exceeds likely deployment costs by almost 2 orders of magnitude. The 2% goal, incidentally, is based on the observation that 5% of smokers will quit following a brief consultation with their GPs on smoking-cessation (Austoker et al., 1994). Hence, if our system can convince even a small number of smokers to quit, it should easily meet cost-effectiveness goals.

### 3.2 Acceptability to Medical Professionals

Most TPI systems are formally evaluated in terms of their impact on *patients*. However, no TPI system is going to be used in the real-world unless it is also acceptable to doctors and other health-care practitioners.

In particular, one issue that comes up in both of our systems is whether individual doctors (or other medical practitioners) perceive enough benefit from the systems to make it worth their while to go through the effort of installing and learning how to use the system. An issue here is that although many younger doctors in Scotland enjoy using computers and are quite keen to try new computer tools, some older doctors are less enthusiastic about using computer-based systems, unless they provide very clear and tangible benefits. Of course, the percentage of "computer-friendly" doctors should increase over time, as the older generation of pre-computer doctors retire.

As mentioned in Section 2.1, this was a major concern with GRASSIC; since using GRASSIC would only result in a small reduction in the number of patients each doctor sent to hospital, there were real doubts as to whether doctors would be willing to make the personal investment required of their time and energy to use the system. Furthermore, using GRASSIC required a significant change in the way doctors interacted with severe asthma patients. The alternative approach of manually customising (improved) booklets, in contrast, did not require doctors to learn new computer skills, and fit much more naturally into the existing procedures for managing asthma patients.

The attitude of doctors as again an issue in smoking-cessation. For instance, as mentioned above, research shows that brief consultations with GPs will help 5% of smokers quit; but yet few GPs regularly make such consultations. This is largely because from a GP's perspective, it is hard to remain excited and motivated about a technique that has a 95% failure rate. This is one of the reasons why we believe it is sensible to try to automate this advice-giving in a computer letter-generator; computers, unlike people, do not get discouraged if they fail in 95% or even 99.99% of cases.

Of course, there is a real possibility that doctors will be reluctant to make the effort to install our letter-generation system. After all, even if it is successful in achieving a 2% cessation rate, from the point of view of an individual medical practitioner, this translates into a very small reduction in the number of his or her patients who smoke. Partially for this reason, we are exploring a number of alternative fielding possibilities for our system, including through GP offices, via health promotion services (such as telephone helplines), inside hospital clinics, and as a service provided by employers. Again it is very early days in our project, but we hope that by exploring several fielding possibilities, we can find one where there is maximal willingness to use our system.

Finally, a fairly obvious point is that individuals will be most willing to use a TPI system if the benefits of the system accrue to them as well as to the health service as a whole. For example, GPs will probably be more willing to use our smoking-letters system if the health service rewards them for lowering smoking rates, or gives them smoking-cessation targets. From this perspective, incidentally, it may well be that most acceptable medical application of NLG is not TPI, but rather systems which help doctors author routine documents (discharge summaries, for example); in such cases the benefits to the individual using the system are much clearer.

### 3.3 Amount of Information Needed

Another important issue for TPI systems is the amount of information they need about patients in order to successfully tailor the documents, and whether this information can be extracted from existing data sources, such as Patient Record System (PRS) databases, or whether it needs to be entered just for the TPI system. A TPI system which requires no extra data will probably be more acceptable to users, since they do not have to spend any time entering information in order to use the system. Similar observations have been made in other NLG projects, e.g., (Reiter et al., 1995).

The GRASSIC system obtained all its patient information from a PRS system; it did not need to acquire additional information for tailoring purposes. However, the PRS system used in the clinic where GRASSIC was developed was relatively advanced. It is not clear whether PRS systems in other Scottish clinics would also contain sufficient information to support GRASSIC's tailoring. Also, the fact that different sites use different PRS systems increases the complexity of installing GRASSIC in a site.

The smoking-letters system, in contrast, requires extensive information to be entered for tailoring; patients must fill out a 4-page questionnaire about their attitudes towards smoking before the system can be used. We are trying to develop ways to make questionnaire entry as easy (and as error-proof) as possible, but the need to enter this information is a significant cost to using the system. On the other hand, because the smoking-letters system makes minimal use of PRS data, it does not need to be customised to the specifics of each site's PRS system, and hence will have a lower installation cost.

The amount of patient-information available to TPI systems should increase over time, as PRS systems become both more comprehensive and more standardised.

### 3.4 Risk and Impact of Mistakes

It is probably inevitable that documents produced by a real-world TPI system will sometimes contain mistakes. This may be a consequence of problems in the tailoring data (for example, incorrect PRS entries or patient questionnaires); it may also be a consequence of bugs in the software.

In some cases mistakes may not be important. For example, if mistakes slightly reduce the effectiveness (via inappropriate tailoring) of a letter encouraging smoking cessation, this is acceptable as long as the TPI system still has sufficient overall effectiveness. If mistakes can lead to medically harmful advice, however, this is a serious problem.

There are a number of solutions to this problem, none of them ideal. These include

- Documents can be reviewed by doctor or nurse before being sent to patients; this was the procedure used in GRASSIC. This may significantly decrease the attractiveness of the system, if the amount of doctor-time required is non-trivial. Human review may not be possible for interactive hypertext systems, such as Migraine (Buchanan et al., 1995) or Piglet (Cawsey et al., 1995), which generate texts "on demand" when requested by patients.
- The TPI system can include disclaimers and warnings in its texts. For instance, tailored nutritional advice which includes recipes (Campbell et al., 1994) could also include warnings such as *do not use this recipe if you are allergic to dairy products*. Such disclaimers will significantly reduce the "personalised" aspect of the generated texts, however, which is the whole purpose of TPI systems.
- The TPI system may be designed so that documents do not contain specific advice or instructions. For example, the smoking-letters system could stress facts (e.g., *have you realised that you are spending £100 per month on smoking*) or motivational stories (e.g., *Many other single mothers have managed to quit. For example Jane Doe...*) instead of advice (e.g., *Start jogging to lose weight*). Of course, if the TPI system is communicating a treatment regime (medication, diet change, etc.), then this approach will not be possible.

We have not yet decided which of the above approaches to use in our smoking-cessation system.

Another possibility is to simply accept that mistakes will occur. Doctors, after all, occasionally make mistakes, and perhaps the right goal for computer systems is not "be perfect" but rather "make mistakes less often than doctors". However, in current medical contexts, computer systems are held to a much higher standard than doctors. If a doctor gives bad advice that causes a patient to become sick, this is regrettable but hardly news. However, if a computer system does the same, even on just one patient out of thousands, it may cause the system to be withdrawn from service.

## 4 Conclusions

TPI systems are likely to be of increasing interest to health care providers. They clearly work to some degree, and they should become more effective as they start using more advanced technology, such as NLG. However, it is not sufficient for a TPI system to be clinically effective in order to be fieldable; it also needs to be cost-effective, acceptable to individual users (patients as well as medical practitioners), have low data-entry costs, and incorporate a satisfactory solution to the mistakes issue. This is a daunting set of requirements, and may explain why although many TPI systems have been developed in the lab, few have been fielded.

We hope that a better understanding of these issues will help TPI developers (including ourselves) produce systems that are more likely to be deployed and used in the real world. We believe that TPI technology has the potential to make a real impact on health, especially given the increasing importance of life-style and compliance issues; good health is mostly a function of actions and decisions taken by patients, not by health-care professionals. But this potential will only be realised if we can build systems that are not only technologically ingenious and clinically effective, but also are easy to deploy and use.

We would like to conclude by saying that we believe that these fielding problems will decrease in the future. In particular, cost-effectiveness should increase as technology improves; acceptance among health-care professionals should become easier as more such people become computer literate and friendly; data-entry should become less of a problem as PRS systems become richer and more standardised; and people may become more tolerant of computer mistakes if they adopt the "make mistakes less often than a doctor" criteria. So, in ten years time it should be much easier to deploy a TPI system; all the more reason for researchers to work today on developing appropriate technology and identifying good applications.

## References

J. Austoker, D. Sanders, and G. Fowler. 1994. Smoking and cancer: smoking cessation. *British Medical Journal* **308**:1478-1482.

B. Buchanan, J. Moore, D. Forsythe, G. Carenini, G. Banks and S. Ohlsson. 1995. An intelligent interactive system for delivering individualized information to patients. *Artificial Intelligence in Medicine* **7**:117-154.

M. Campbell, B. DeVellis, V. Strecher, A. Ammerman, R. DeVellis, and R. Sandler. 1994. Improving dietary behavior: the effectiveness of tailored messages in primary care settings. *American Journal of Public Health* **84**:783-787.

A. Cawsey, K. Binsted and R. Jones. 1995. Personalised explanations for patient education. In *Proceeding of the 5th European Workshop on Natural Language Generation*, pages 59-74, Leiden, The Netherlands.

A. Cawsey and F. Grasso. 1996. Goals and attitude change in generation: a case study in health education. In *Proceedings of the ECAI 96 Workshop "Gaps and bridges: new directions in planning and natural language generation*, pages 19-24, Budapest, Hungary.

A. Lahdensuo, T. Haahtela, J. Herrala, T. Kava, K. Kiviranta, P. Kuusisto, E. Perämäki, T. Poussa, S. Saarelainen, and T. Svahn. 1996. Randomised comparison of guided self management and traditional treatment of asthma over one year. *British Medical Journal* **312**:748-752.

L. Osman, M. Abdalla, J. Beattie, S. Ross, I. Russell, J. Friend, J. Legge, and J. Douglas. 1994. Reducing hospital admissions through computer supported education for asthma patients. *British Medical Journal* **308**:568-571.

E. Reiter. 1995. NLG vs. templates. In *Proceedings of the Fifth European Workshop on Natural Language Generation*, pages 95—105, Leiden, The Netherlands,.

E. Reiter, C. Mellish, and J. Levine. 1995. Automatic generation of technical documentation. *Applied Artificial Intelligence* **9**:259-287.

E. Reiter, A. Cawsey, L. Osman, and Y. Roff. 1997. Knowledge acquisition for content selection. In *Proceedings of the 1997 European Workshop on Natural-Language Generation,* pages 117-126, Duisberg, Germany.

W. Velicer, J. Prochaska, J. Bellis, C. DiClemente, J. Rossi, J. Fava, and J. Steiger. 1993. An expert system intervention for smoking cessation. *Addictive Behaviors* **18**:269-290.